\documentclass[letterpaper, 10 pt, conference]{ieeeconf}  % Comment this line out if you need a4paper

\IEEEoverridecommandlockouts                              % This command is only needed if 
\usepackage{cite}
\usepackage{amsmath,amssymb,amsfonts}
\usepackage{algorithmic}
\usepackage{graphicx}

\usepackage{caption}
\usepackage{subcaption}
\usepackage{textcomp}
\usepackage{xspace}
\usepackage{balance}
\usepackage{paralist}
\usepackage{url}
\usepackage[bottom]{footmisc}
\usepackage{hyperref}
\usepackage[dvipsnames]{xcolor}
\usepackage{tabularx} % for 'tabularx' environment
\usepackage{multirow}

\usepackage{latexsym}         % Fuer recht seltene Zeichen
\usepackage{makecell}
\usepackage{booktabs}

\title{\LARGE \bf
Safe Perception - A Hierarchical Monitor Approach
}

\author{Cornelius Buerkle$^{*}$, Fabian Oboril$^{*}$, Johannes Burr and Kay-Ulrich Scholl$^{*}$
\thanks{$^{*}$Intel Labs, Karlsruhe, Baden-W\"urttemberg, Germany. Emails: \url{ { cornelius.buerkle,fabian.oboril,kay-ulrich.scholl } @intel.com}}
}

\begin{document}
\bstctlcite{IEEEexample:BSTcontrol}

\maketitle
\thispagestyle{empty}
\pagestyle{empty}

%%%%%%%%%%%%%%%%%%%%%%%%%%%%%%%%%%%%%%%%%%%%%%%%%%%%%%%%%%%%%%%%%%%%%%%%%%%%%%%%
\begin{abstract}
Our transportation world is rapidly transforming induced by an ever increasing level of autonomy. However, to obtain license of fully automated vehicles for widespread public use, it is necessary to assure safety of the entire system, which is still a challenge. This holds in particular for AI-based perception systems that have to handle a diversity of environmental conditions and road users, and at the same time should robustly detect all safety relevant objects (i.e no detection misses should occur). Yet, limited training and validation data make a proof of fault-free operation hardly achievable, as the perception system might be exposed to new, yet unknown objects or conditions on public roads. Hence, new safety approaches for AI-based perception systems are required. For this reason we propose in this paper a novel hierarchical monitoring approach that is able to validate the object list from a primary perception system, can reliably detect detection misses, and at the same time has a very low false alarm rate.
\end{abstract}

%%%%%%%%%%%%%%%%%%%%%%%%%%%%%%%%%%%%%%%%%%%%%%%%%%%%%%%%%%%%%%%%%%%%%%%%%%%%%%%%
\section{Introduction}

The development of Automated Vehicles (AVs) has made great progress over the last years. As a result, more and more prototypes are tested on public roads and the first robotaxi services have been launched as well~\cite{mobileyeMunich2020, reportAV2020CA, waymo2020Robotaxi}. Despite this success, there are still many open challenges demanding attention to allow mass deployment, in particular with respect to safety assurance under all possible  conditions. %In fact, without comprehensive safety systems in place, no AV will receive certification.

In this regard, a special focus has to be put on the perception systems. First, contrary to the decision making aspects of an Automated Driving System (ADS) for which an IEEE standard is about to be published ~\cite{ieeeAVSafety}, comprehensive safety concepts for perception systems are still missing. At the same time, any perception error, for example a not-detected object, may propagate through the AV processing pipeline and result in a dangerous driving behavior, even if safety concepts for the decision making are in place. In addition, despite the great progress in the past years on AI-based object detection, it is still impossible to prove that these systems robustly detect all objects under all possible conditions (night, rain,  highway, urban, etc.)~\cite{nuScenesLeaderboard,DBLP:KIAbsicherung}. In fact, generalization and robust detection of rare objects (potentially not in training set) remains an open challenge \cite{burton2017making}.

\begin{figure}[t!]
\centering
\includegraphics[clip,trim=0.0cm 9.5cm 16.cm 5.cm,width=0.9\columnwidth]{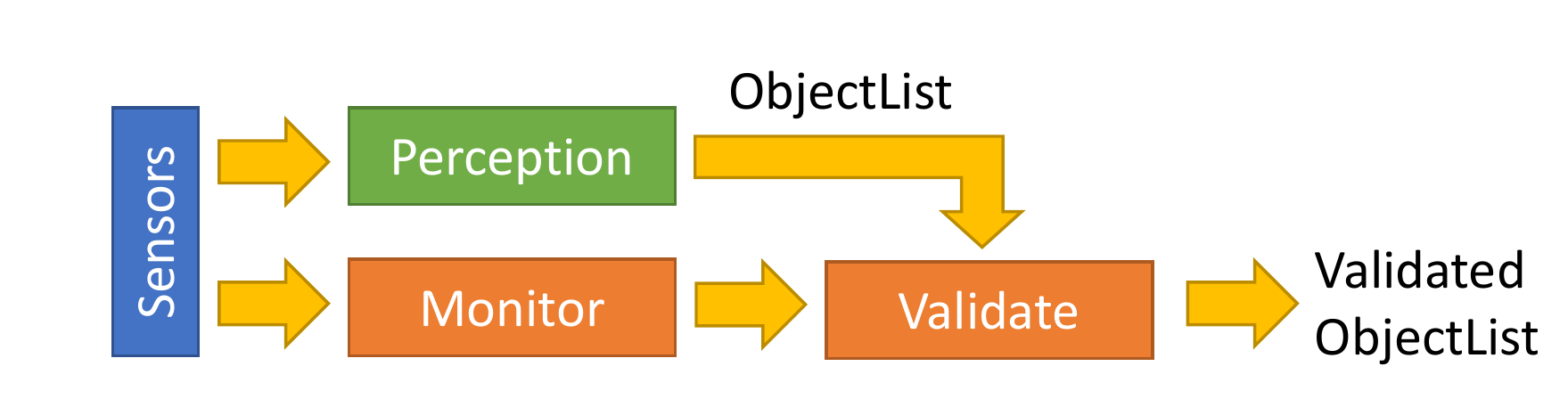}
\vspace{-0.15cm}
\caption{Monitor architecture}
\label{fig:MonitorIdea}
\vspace{-0.6cm}
\end{figure}

Consequently, there is a strong need for verifiable safety systems for AV perception. In addition, these should ideally be computationally lightweight to be able to run on safety-certified hardware which is typically less powerful. To achieve this goal, it is often of an advantage to build a monitor architecture that only verifies the correctness of the results of the primary perception system (see Figure~\ref{fig:MonitorIdea}), instead of developing means to assure safety of the primary perception system. This is similar to verifying that a result of a complex mathematical equation is correct, which is usually simpler than deriving the result itself. A first proposal of such a perception monitor architecture was published in~\cite{buerkle2020onlineenv}, where the monitor uses LiDAR sensor data to create a dynamic occupancy grid. This concept was extended in~\cite{BuerkleFaultTolerant} with additional plausibility checking. While the monitor was able to detect and correct detection misses in the primary perception system, it is still pretty computationally intensive. In addition, in both studies it was assumed that LiDAR noise, detection points related to the road surface, etc. can easily be excluded when creating an occupancy grid without safety implications. In practice, this is however not straightforward.

Another generic challenge of such a monitor architecture is the need to keep the false alarm rate on a very low level, while robustly detecting all objects that were missed by the primary channel. Typically, detection misses (false negatives) and false alarms (false positives) are inversely correlated, i.e. if the monitor has a low detection miss rate, it will suffer from a high false alarm rate. This however, will hamper the availability (practicality) of the entire system as a high false alarm rate generated by the monitor can lead to frequent, undesired braking maneuvers by the driving policy.

To overcome this challenge, we propose in this paper a novel perception monitor system illustrated in Figure~\ref{fig:Concept}. It uses a hierarchical filter concept, built around a model-based height confidence filter, which ensures that everything that is likely a safety-relevant object is detected and that everything that belongs to the road surface is ignored. Additional filters are then added to evaluate ambiguous detections. This creates a filtered, sparse point cloud that is then converted into an occupancy grid. As a result, our envisioned monitor can achieve a low false negative rate ($<0.2 \%$), while maintaining a low false positive rate($<2\%$). 

The remainder of this paper is organized as follows. In Section~\ref{sec:SafetyRelevance}, we introduce the problem statement and formulate the notion of safety relevance. Based on that, the monitor concept is motivated in Section~\ref{sec:concept}, followed by an in-depth explanation in Section~\ref{sec:Realization}. Afterwards, an experimental study is presented in Section~\ref{sec:Evaluation}, and finally the paper is concluded in Section~\ref{sec:conclusion}.

\section{Problem statement}
\label{sec:SafetyRelevance}

With this work we want to demonstrate an approach that allows a system to detect all safety relevant objects in the surroundings of the vehicle. To level set these goals we will start with a  definition of the underlying assumptions.

\subsection{Safety Relevance}
First of all not all objects in the surrounding of an AV are safety relevant. For example, objects that are further away or driving with a high positive velocity delta can be considered as not relevant, as a collision with such an object is not possible in near future. Consequently, it is possible to define a safety relevant zone in which all relevant objects are located~\cite{SafetyRelevance2021}. In addition, only the closest object of each lane can have a direct impact on the AV. Thus, the design goal is to have a system in which at least the closest object on every lane within the safety relevant region is detected.

\subsection{Monitor}
Our goal in this work is to develop and propose a novel monitor for safe perception as depicted in Figure \ref{fig:MonitorIdea}. In other words, our intent is to create a monitor component that operates alongside an existing (AI-based) perception system. The combination of perception system and monitor, shall be robust enough that the combined perception system doesn't miss a safety-relevant object.
Hence, the focus of the monitor is to identify, if there are objects in the environment that have been not detected. To achieve this, the monitor should be able to detect the presence of all objects in the safety relevant region, while an accurate determination of dimension or bounding box is not required or in focus. In addition, a low false alarm rate is a second objective for the monitor.

\subsection{Sensor Information}
The primary sensor information for our monitor is a LiDAR point cloud (PCL). It is assumed that the safety relevant objects are visible to the vehicle. If there are occlusions within the field of view, these can be handled by different means \cite{shalev2017formal}, but are not in scope for the current work. 
It is further assumed, that the LiDAR provides measurements for all non-occluded safety relevant objects.     

To prove this point we analyzed the LiDAR information and annotations of the Lyft~\cite{LyftDS} dataset. 
Therefore we checked for each annotation how many LiDAR measurement points are within the 3D bounding box of the object. Our analysis shows that the majority of relevant objects ($>99.8\%$) contains more than two measurement points and therefore should be detectable by the perception system. The few objects without any measurements are all totally occluded.

%Besides there has been great progress in the detection quality of perception systems in the last years. Still even the best performing networks ain't perfect yet and miss to detect relevant objects.

%At least for the vision benchmarks, this miss detection are mainly caused by imperfections of the perception algorithms.
%Therefore these vehicles are 

%This shows that all annotated objects are detected by LiDAR system. For the dataset this might be not a real surprise, as the annotations are most likely created based on the information of the LiDAR point cloud. Nevertheless it proofs that the perception algorithm fail to identify all objects in the LiDAR information.

%\subsection{Goal}
%As all relevant objects are visible in the LiDAR information we want to use this information to assure a correct perception.
%Our goal is to construct a monitor for a perception system based on LiDAR information.
%The monitor should reliably detect miss detections of the primary perception system by keeping a low false alarm rate. 
%
%Besides that the monitor should guarantee that no objects are overseen that are clearly visible in the LiDAR point cloud, regardless of their shape.

\section{Concept}
\label{sec:concept}

In our previous work~\cite{buerkle2020onlineenv,BuerkleFaultTolerant} we described how an object list provided by a primary perception system can be verified using an occupancy grid. The fundamental principle is that the object position and bounding box information can be converted into a spacial occupancy probability. This occupancy probability can be correlated with the occupancy grid provided by a secondary perception system. Assuming this secondary occupancy information is correct, errors in the primary object information can be identified by calculating the conflict of the two different spatial occupancy values.

\begin{figure}[t!]
\centering
\includegraphics[clip,trim=0.2cm 7.cm 1.cm 0.2cm,width=0.95\columnwidth]{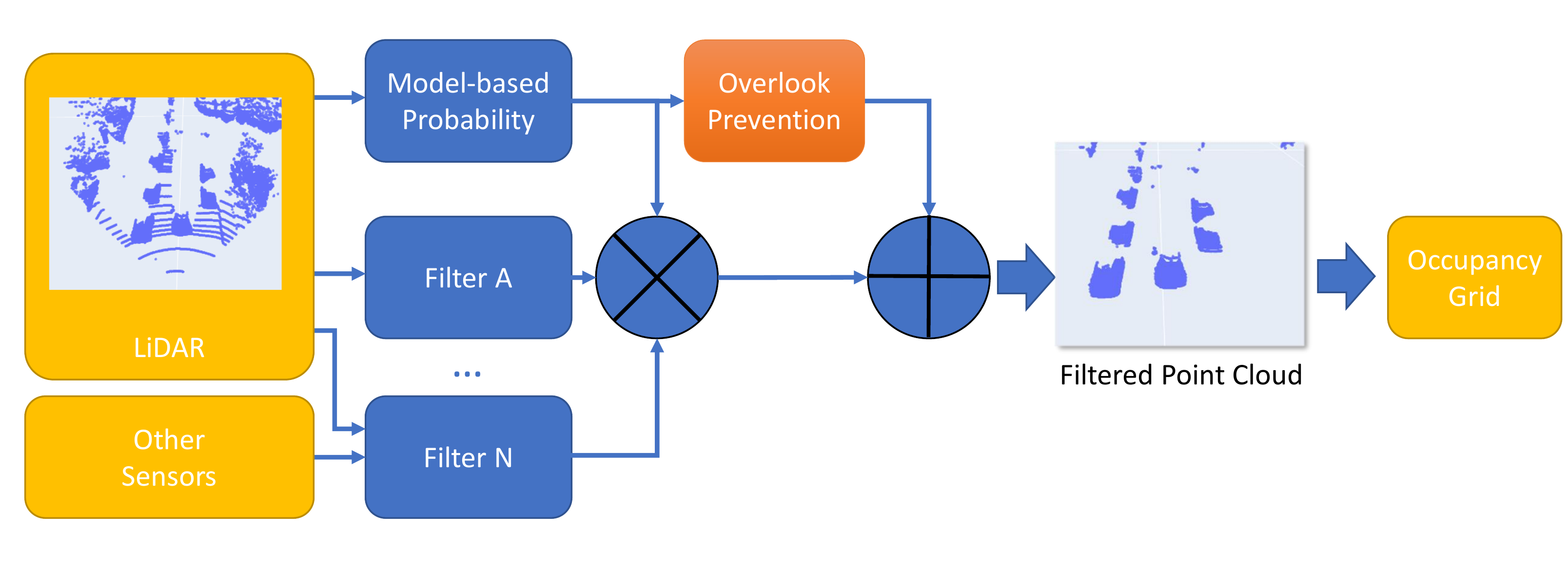}
\vspace{-0.15cm}
\caption{Hierarchical Monitor Concept with Overlook Prevention, Filtered Point Cloud and resulting Grid}
\label{fig:Concept}
\vspace{-0.45cm}
\end{figure}

To ensure safety and availability of the overall system, the secondary perception channel needs to fulfil two key requirements. First, all obstacles have to be included in the occupancy grid (to be safe), and second, driveable space should not be marked as occupied (otherwise false alarms will reduce the availability). However, when using a LiDAR  sensor for the secondary channel, by default drivable space will be included in the point cloud. Hence, a comprehensive, safety-aware filtering of the LiDAR point cloud is required to avoid that drivable space is flagged as occupied region, before it can be converted into an occupancy grid.

In this work, we present a novel solution on how such a secondary perception system (a.k.a monitor) can be constructed, that a) uses LiDAR sensor data as input (as all relevant objects are typically represented within LiDAR data as explained before), b) is able to detect all safety relevant objects that may have been missed by the primary perception system, c) has a low false-alarm rate and d) is robust to varying road surfaces and LiDAR noise levels.

For this reason, we propose a hierarchical filter concept to process the LiDAR point cloud, as depicted in Figure~\ref{fig:Concept}. Each of the filters determines for each point within the point cloud a probability that the point is a measurement of a relevant obstacle or object. The filters can therefore make use of the LiDAR information, but can also take other sensor information into account. By combination of these probabilities we are able to create a robust overall decision for each point whether it belongs to a relevant obstacle or if it is traversable (AV can safely drive over or under it). This allows to filter out all irrelevant points and finally create an occupancy grid from the filtered point cloud.

%As described our working assumption is that all relevant objects are visible in the LiDAR point cloud. Therefore the purpose of our system is to reliably identify regions in the point cloud that correlate with relevant objects and remove points that are not relevant for collision avoidance, like road surface or noise.  

In this regard it is important that the point filtering method safeguards against erroneous deletion of points that belong to a relevant object (otherwise safety is impaired), yet at the same time points related to drivable space should be eliminated. Unfortunately, very often information is ambiguous and it is not directly clear to which category a point belongs to. To create a system in which objects that are clearly visible within the point cloud are not missed by the perception system, an "overlook prevention" component is created. This component checks that all points with high probability of belonging to a relevant object are included in the final result. To handle ambiguous detection points and reduce the chance of false alarms, additional filters are used to specifically evaluate these ambiguous points. In other words, the additional filters will never remove points that are flagged as relevant by the ``overlook prevention'' component, but may flag additional points as relevant. As a result, the combination of the other filters will increase the accuracy and reliability in regions where the LiDAR information is more ambiguous.

\section{Realization}
\label{sec:Realization}
\subsection{Model-based Probability Filter}

The first component of the proposed hierarchical perception monitor is a model-based probability filter, which uses model knowledge to determine the probability for each point in the point cloud of belonging to a relevant object. For this reason, this filter uses height and depth information, as we will explain next.

When looking at driving scenarios we can categorize any surface or object in the surrounding of the vehicle that creates a measurement in the LiDAR point cloud into two categories:

\begin{enumerate}
 \item[Traversable:] These are objects and surfaces that the car can traverse over or under without any harm. Examples are road surfaces or speed bumps on the one hand, or bridges or branches of trees on the other hand.    
 \item[Obstacles:] Any other object in the environment. As the AV cannot bypass these without causing a collision.
 \end{enumerate} 
 
The goal of the model-based probability filter is to reliably identify measurements that belong to obstacles. Such a probability can be defined by applying some obvious model knowledge. 
An object or surface point is traversable if it belongs to the road surface or its height above the road surface is large enough for the vehicle to drive through, we regard this as ceiling.
Anything not falling into these two categories can be an obstacle. 
Unfortunately it is not straight forward to determine if an object is traversable.
A road surface might not be flat, the road might have an incline or there might be small obstacles on the street that could be dangerous to the car. 
But it is possible to at least identify parts of the environment that have a high potential of being an obstacle $P_O$. Briefly explained, these are all parts of the environment that have a significant height over the road surface, or show a significant height incline.

\subsubsection{Preprocessing}
~\newline
\indent In order to calculate the described  $P_O$ we need to be able to determine height differences between adjacent measurement points. Unfortunately this is not directly possible for an (unsorted) LiDAR point cloud, as neighbor relations among points cannot be inferred. 
To overcome this issue we convert the point cloud into a height and depth image, where each pixel corresponds to exactly one LiDAR measurement point. Thus, the height of these images is equivalent to the number of vertical LiDAR scans (layers), and the image width is given by the horizontal resolution of each LiDAR scan. Given the LiDAR point positions in 3D coordinates within the sensor coordinate system - $(x,y,z)$ , $z$ describing the height and $x$ and $y$ distance to front and left -  we need to calculate the azimuth $\alpha$ and vertical angle $\delta$ , as follows:  

\begin{eqnarray}
\alpha = \arctan(y,x) \label{eqn::pclmatrix1} \\
\delta = \arctan(z, \sqrt{(x^2+y^2)})
\label{eqn::pclmatrix2}
\end{eqnarray}

Given these values,  $\alpha$ and  $\delta$, it is possible to assign each point to exactly one pixel and therefore a lossless conversion of the point cloud is maintained. Unfortunately noise on point positions and the movement of the vehicle during the duration of the scan can introduce errors. These errors can cause several points to be associated with one pixel. For these conflicts we discard, the point value that has a higher distance. Using that approach we can guarantee that the minimum distance towards a measurement point is always maintained. Of course, if the sensor provides already the output in form of a row-column matrix, this step is not required and the height and depth images can be directly inferred from the sensor data.

\begin{figure}[t!]
\centering
\includegraphics[clip,trim=1.0cm 6.cm 6.cm 3.cm,width=0.95\columnwidth]{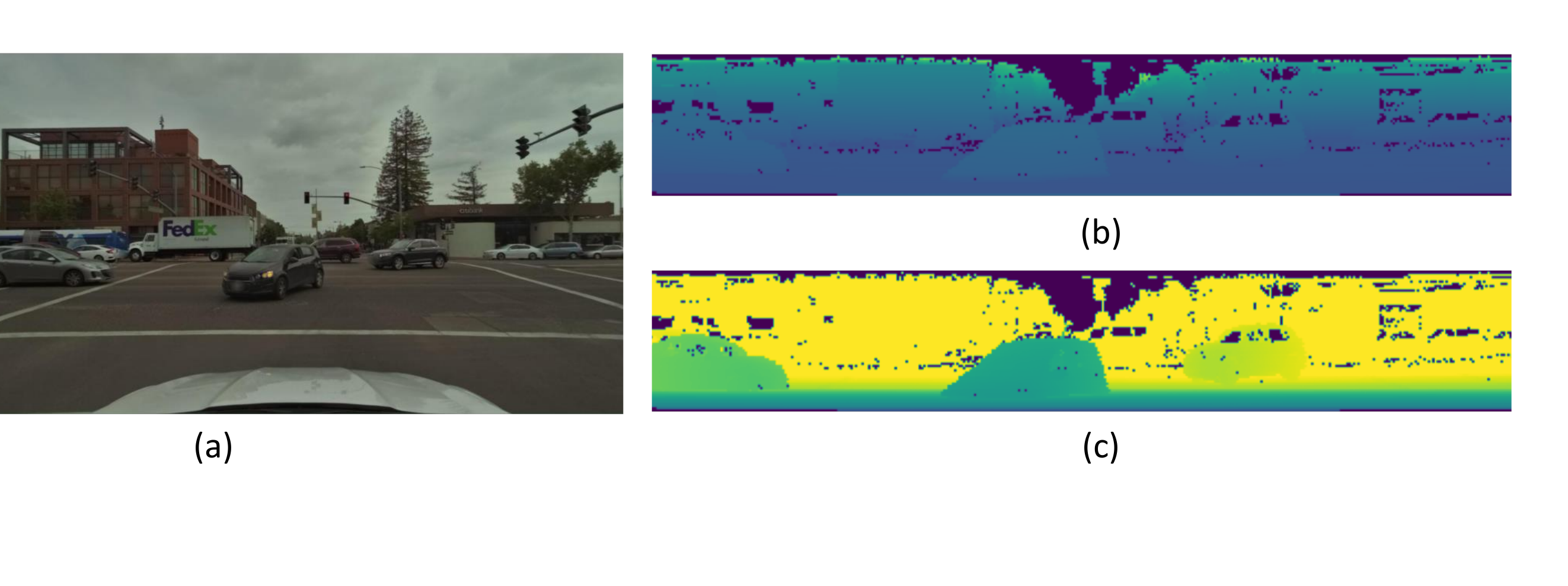}
\vspace{-0.55cm}
\caption{(a) Camera image (b) height image (c) depth image}
\label{fig:pclImage}
\vspace{-0.65cm}
\end{figure}

Once the pixel position is calculated we construct two images, as illustrated in Figure~\ref{fig:pclImage}. A depth image where pixels store the distance to the measurement point and a height image where each pixel provides the height value. For depth and height values we use the vehicle coordinate system as reference, as this allows easier processing of the values in the following processing steps.

\begin{figure}[b!]
\centering
\vspace{-0.25cm}
\includegraphics[width=0.95\columnwidth]{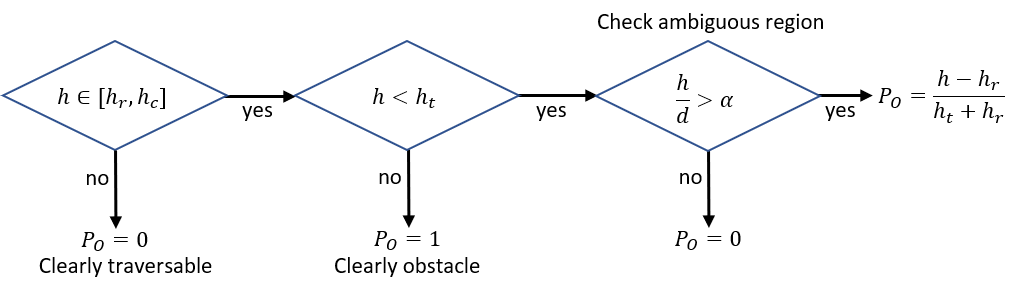}
%\vspace{-0.55cm}
\caption{Steps to estimate potential obstacle score using height and depth information, where $h$ is the current height of the point, $d$ is the distance, $\alpha$ an incline threshold value, $h_r$ and $h_c$ are the road surface and ceiling height, and $h_t$ is a height threshold value}
\label{fig:heightConfidence}
%\vspace{-0.65cm}
\end{figure}

\subsubsection{Obstacle score calculation}
~\newline
\indent The described depth and height image are used to calculate the potential obstacle score $P_O$, according to the steps illustrated in Figure~\ref{fig:heightConfidence}.
$P_O$ should be a low value for all points that are either part of the road surface or the ceiling. The score should increase the further the point is away from road surface or ceiling. 

\begin{figure}[t!]
\centering
\vspace{-0.35cm}
\includegraphics[clip,trim=16.0cm 6.cm 0.cm 3.cm,width=0.75\columnwidth]{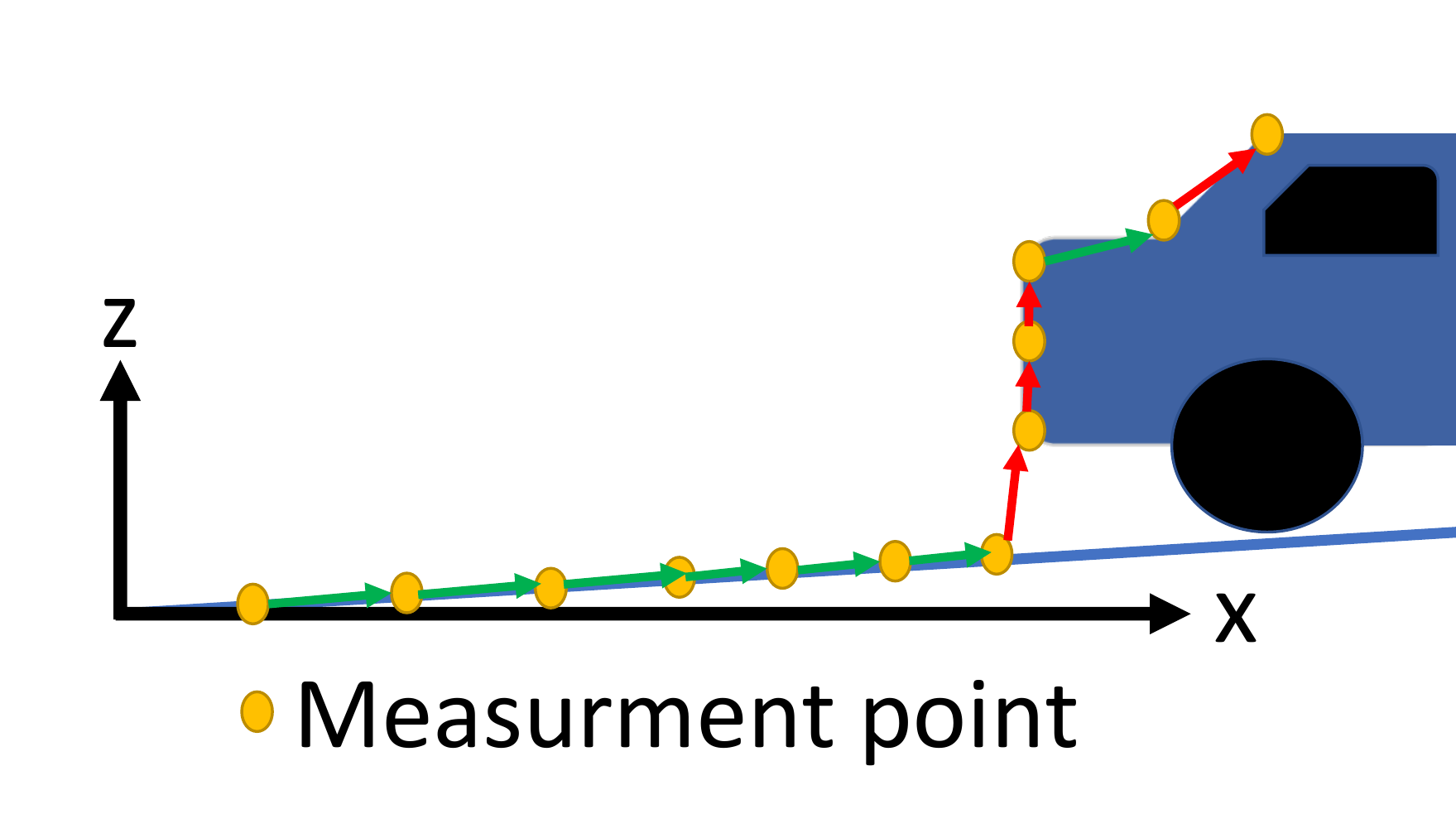}
\vspace{-0.15cm}
\caption{Incline between measurement points. On road surface incline is small. Obstacles and objects lead to strong incline.}
\label{fig:surface}
\vspace{-0.45cm}
\end{figure}

%@TODO provide formulas for the algorithm
Hence, first of all the road surface height needs to be estimated from the LiDAR data, which can be achieved using the height image. The image is constructed in a way that the lowest LiDAR beam correlates with the lowest row in the image. Thus, the road surface should start in the lowest image rows and its height value should be close to zero.
By traversing from bottom to top, the incline between adjacent measurement points can be calculated. As long as the measurements are on the road surface the incline is low, while its exact value will depend on the slope of the road and pitch of the vehicle. In our evaluations a maximum incline of $\alpha=5^{\circ}$ provided robust results. Measurements on obstacles show a way greater incline, as illustrated in Figure~\ref{fig:surface}. Once an obstacle is identified maximum road surface height can be estimated by the height value of the last measurement.

\begin{figure}[b!]
\centering
\vspace{-0.35cm}
\includegraphics[clip,trim=0.0cm 0.cm 0.cm 0.cm,width=0.95\columnwidth]{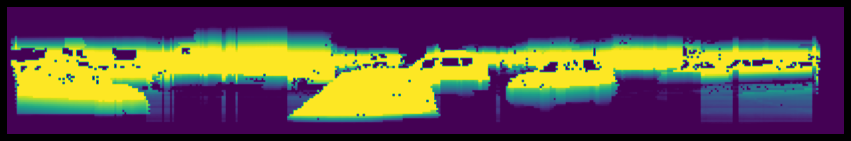}
\vspace{-0.15cm}
\caption{Image of object score. Cars have high object score. Road surface and lower parts of cars have smaller values.}
\label{fig:op}
\end{figure}

With the known road surface height $h_r$ and a minimum ceiling height $h_c$, we can define potential obstacle score $P_O$ as illustrated in Figure~\ref{fig:heightConfidence}. As a result, all points outside of $[h_r, h_c]$ (which are traversable) have $P_O = 0$, for all points that have a height that is considered as relevant (i.e. $h\in[h_t, h_c]$) $P_O=1$, and for the ambiguous points the incline is evaluated and $P_O$ is set according to Figure~\ref{fig:heightConfidence}. With this definition obstacles and objects that are clearly visible in the LiDAR measurements will generate a high $P_O$, as shown in Figure~\ref{fig:op}.

%Therefore we can use this score to construct our high confidence filter. We just need to assure that measurements with high $P_O$ will get an high overall probability. 

This filter helps the system to not miss clearly visible obstacles. Nevertheless, there are regions where a decision based on $P_O$ score is not obvious.
These could be partially occluded objects, stronger slopes of the road and many more.
In this work we used some trained perception algorithms as additional sources of information to come to a better decision whether these regions are relevant or not. It might be argued that this will weaken the safety argument, as AI algorithms bring the risk of miss classification due to incomplete training data and lack of generalization. 
To counter this problem we propose a methodology to integrate the information in a way, that the AI information cannot jeopardize the safety of the system.

In our work we used two additional perception algorithms which we will describe now.
It is worth mentioning that the architecture allows to integrate additional filters, beyond the ones presented next.

\subsection{Objectiveness}

As a second source of information we estimate a likelihood of belonging to an object $P_N$ using an AI algorithm trained on LiDAR information. 

The used neural network architecture is strongly inspired by PointPillars ~\cite{PointPillars}, which was one of the first models providing good 3D bounding box detection out of LiDAR information, by maintaining an efficient real time calculation. Our architecture is further simplified by exchanging the detection head by a convolutions structure outputting a 512x512 2D grid. We are calling the activation of each cell the objectiveness score indicating the likelihood the model assigns that the given cell is occupied by an object or not. 

\subsubsection{Training}

For training, the dataset was split into a training and validation set. The range of the point cloud was constrained to $\pm 81.92m$ around the ego vehicle. The output is a 512x512 2D-grid, where each cell can either be occupied or non-occupied by an object. The network output is transformed by a sigmoid activation into the range $[0;1]$. We are calling this value the objectiveness score of this grid cell, which can be interpreted as the likelihood the network assigns to it being occupied.
The given 3D annotations were transformed to 2D areas discretized to grid cells of the same size. The model was trained via a Binary Cross Entropy Loss. For each output cell $i$ the activation $\hat{y}$ is compared with the ground truth $y$ via equation (\ref{eq:training}).

On a subsample the imbalance of occupied to unoccupied cells was estimated to be roughly 1:50, which was included as a positive weighting $w=50$ in the loss in order to deter the model to only predict background. The loss function conclusively looks like this:
\begin{equation}
\ell_i(y,\hat{y}) = w \cdot y_i \cdot log(\hat{y}_i) + (1-y_i) \cdot log(1-\hat{y_i}) 
\label{eq:training}
\end{equation}

The overall loss for each data point is then calculated as the mean loss $\ell_i$ of all cells.
$$
L = \frac{\sum_{i}^{N}\ell_i}{N}
$$

\subsubsection{Potential object score estimation}

\begin{figure}[t!]
\centering
\includegraphics[clip,trim=0.0cm 0.cm 0.cm 0.cm,width=0.95\columnwidth]{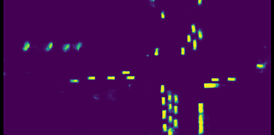}
\vspace{-0.15cm}
\caption{Objectiveness results for region around ego vehicle}
\label{fig:objectiveness}
\vspace{-0.45cm}
\end{figure}

As said, the objectiveness information provides a probability score for a region around the car that this region is occupied by an obstacle [Figure ~\ref{fig:objectiveness}].
With this information estimation of probability to belong to an object $P_N$ for a point in the point cloud is straight forward, by just comparing x and y position and retrieving the probability in the objectiveness grid.

\subsection{Semantic Segmentation}

As a third source of information we use semantic segmentation. For practicality reasons we used camera based semantic segmentation, as this is already publicly available with the OpenVINO SDK. Nevertheless, it is also possible to use LiDAR based semantic segmentation instead to avoid usage of different modalities. 
To realize the filter, we feed the camera images into the OpenVINO FastSeg-Large model, which was built on the MobileNetV3 large backbone and uses a modified segmentation head based on LR-ASPP. More information can be found in~\cite{semSegModel}.

\begin{figure}[t!]
\centering
\includegraphics[clip,trim=0.0cm 0.cm 0.cm 0.cm,width=0.95\columnwidth]{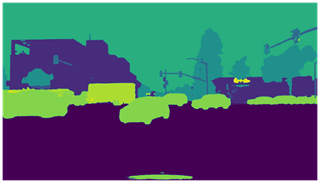}
\vspace{-0.15cm}
\caption{Semantic segmentation image}
\label{fig:semantic}
\vspace{-0.45cm}
\end{figure}

Having the segmented image (see Figure~\ref{fig:semantic}), where each pixel is colored according to the class it belongs to (e.g. car, road, sky, etc.), the LiDAR points are projected onto this segmented image. We assign a probability $P_S$ based on the classification of the pixel $i,j$ associated with the point as well as the classes of the neighboring pixels as follows:

\begin{equation}
\footnotesize
\begin{split}
P_S (i,j) &= 0.5 \times 1_{(i,j)}  \\
          &+ 0.1 \times \left[ 1_{(i-1,j)} + 1_{(i+1,j)} + 1_{(i,j-1)} + 1_{(i,j+1)}\right]  \\
          &+ \frac{1}{40} \times \left[ 1_{(i-1,j-1)} + 1_{(i+1,j-1)} + 1_{(i+1,j-1)} + 1_{(i+1,j+1)}\right]
\end{split}
\end{equation}
where $1_{(i,j)}$ is 1 if the class for this pixel is considered relevant and 0 otherwise. 

By taking not only the pixel itself, but also the neighboring pixels into account, we account for the fact that semantic segmentation is often not very precise on the edges of objects. By applying specific weights based on pixel distance, this challenge can be alleviated. Here sample weights are illustrated that provided the best results in our case, yet for other configurations or resolutions different weights might be required. 

\subsection{Fusion with Overlook Prevention}

For each point in the LiDAR point cloud we now have three scores $P_O$, $P_N$ and $P_S$. Based on these values the point cloud is filtered. 

In order to realize the aforementioned ''Overlook Prevention``, we make sure that points with a high $P_O$ are not filtered out. This is achieved by the first case of Equation~\eqref{eq:combination} 

For points with lower $P_O$ we use a combination of the other values to increase reliability and robustness.
In our experiments filtering according to $f$ showed best results. $f$ will return $1$ for all points $i$ that should remain, $0$ if the point should be removed.

\begin{equation}
\footnotesize
 f(P_O,P_N,P_S) = \\
 \begin{cases}
 \text{1,} & \mbox{if }  P_O > 0.9 \\
 \text{1,} & \mbox{if }  P_O > 0.5 \land P_S > 0.7 \land P_N > 0.2 \\
 \text{1,} & \mbox{if }  P_O > 0.5 \land P_N > 0.6 \\
 \text{1,} & \mbox{if }  P_O > 0.5 \land P_S > 0.9 \\
 \text{0,} & \mbox{otherwise}
\end{cases}
\label{eq:combination}
\end{equation}

\subsection{Monitor}
\label{realization:monitor}
\begin{figure}[t!]
\vspace{+0.15cm}
\centering
\includegraphics[clip,trim=2cm 6.cm 13.cm 2.cm,width=0.95\columnwidth]{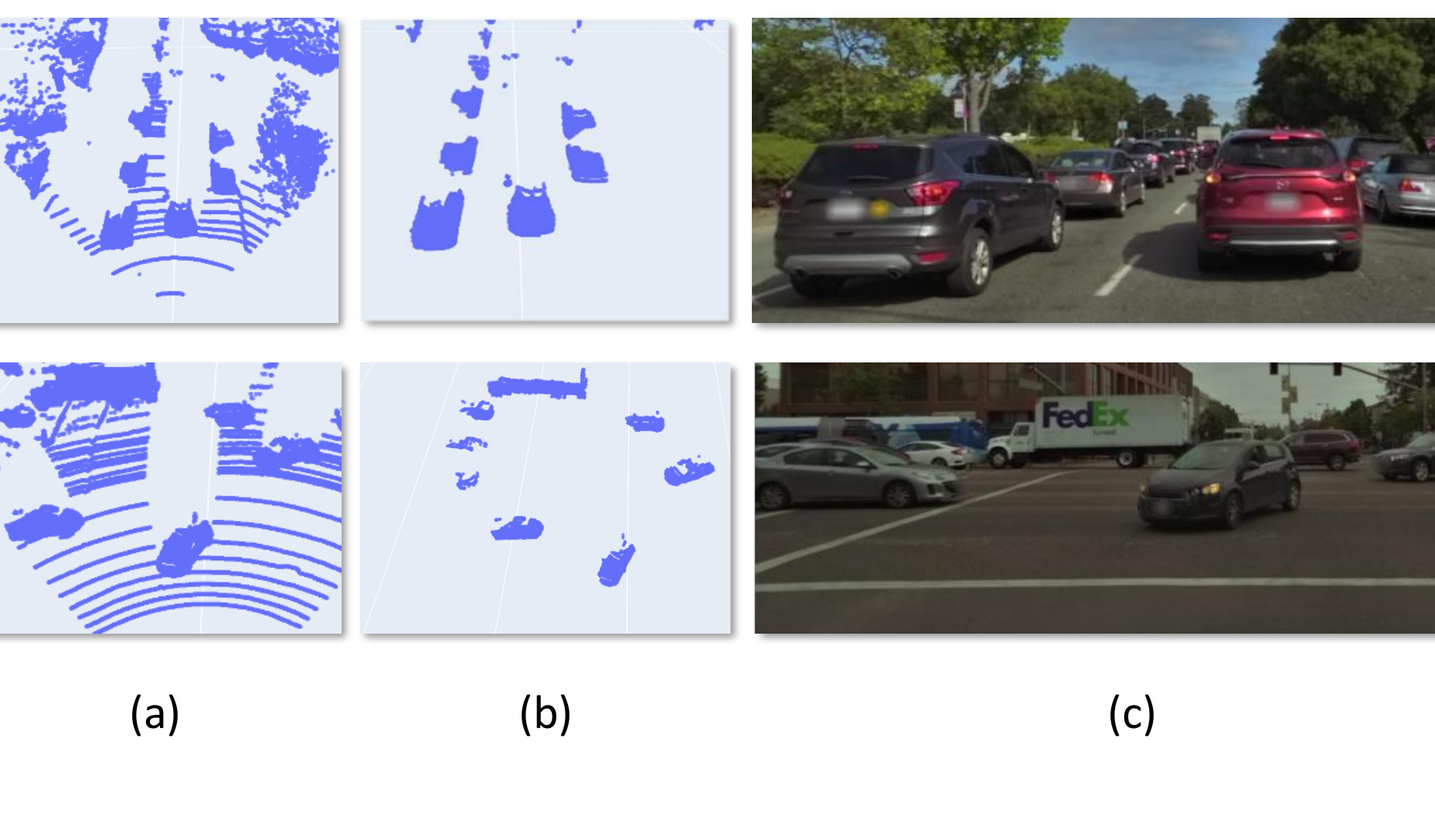}
\vspace{-0.15cm}
\caption{(a) Original point cloud (b) Filtered point cloud (c) Camera image}
\label{fig:filter_pcl}
\vspace{-0.45cm}
\end{figure}

The result of the filtering is a reduced point cloud in which each point has a high confidence to be part of an obstacle [Figure \ref{fig:filter_pcl}]. In order to become robust against noise we convert this point cloud in an occupancy grid as described in \cite{BuerkleFaultTolerant}.

%\vspace{-0.2cm}
\section{Evaluation} 
\label{sec:Evaluation}

In order to demonstrate the feasibility of our approach we run experiments using the Lyft~\cite{LyftDS} dataset. %and KITTI360 ~\cite{Liao2021ARXIV} 

\subsection{Methodology}

For the development and evaluation we split the datasets in a training and validation set, where the validation set covers approximately 5\,\% of all frames of each scene in Lyft.
%[@TODO more details]

The training set was used only for training of the objectiveness and a reference perception system. For the semantic segmentation model a pre-trained version from the OpenVINO model zoo was used. After training, we obtained the results presented afterwards on the validation set.

As described in~\cite{SafetyRelevance2021}, by using a reasonable worst-case assumption of the motion behaviour of other traffic participants, it is possible to define a safety relevant zone around the ego vehicle, in which all objects will be located, that could impose a threat to the ego vehicle.
In other words only objects within the safety relevant zone can have an impact on the vehicle safety.  Therefore, we restrict our evaluation to this zone. According to the definition in~\cite{SafetyRelevance2021} this area covers 36m in front of the vehicle and 8.5m to the side. In addition, we further restrict this area to consider only parts of the environment that belong to the roads.

\subsection {Reference perception system}

To establish a reference for our evaluation we use a state-of-the-art perception algorithm, namely PointPillars~\cite{PointPillars}, which is a neural network for 3D object detection with LiDAR information. The network was trained on the training samples of the datasets. To evaluate the performance, we run inference of the network for the test samples and collected detection rates for the objects within the safety relevant zone. To get a better impression of the performance, we also performed multiple evaluation runs with different confidence thresholds $\tau_{conf}$ for the predicted bounding boxes. As we are primarily interested if the objects have been detected or missed, we use an IOU $>0$ between detection and annotation bounding box to rate a correct detection. The results of our experiments are outlined in TABLE~\ref{table:eval}-[5-9].

Our evaluation shows, that the detection quality of PointPillars is very dependent on the used confidence threshold $\tau_{conf}$. If we use a conservative threshold of 0.5 the numbers of missed objects (FN count) is 90, which is pretty high. When decreasing the threshold to 0.2 the number of missed objects reduces to 20. However, this comes at the cost that the number of false alarms (FPs) increases from 9 to 115. 

\begin{figure}[t!]

\centering
\includegraphics[clip,trim=1.2cm 2.cm 14.cm 5.cm,width=0.95\columnwidth]{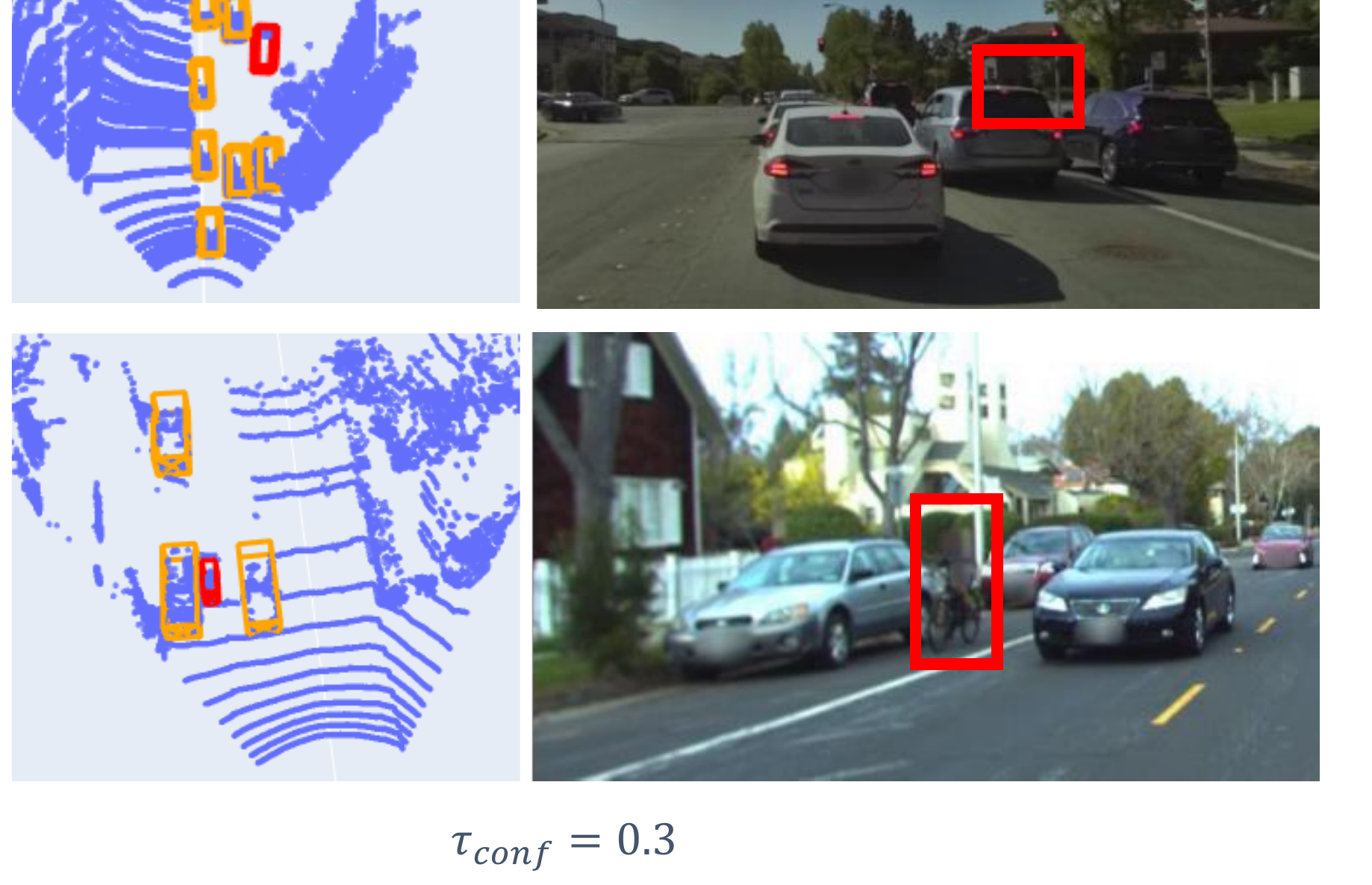}
\vspace{-0.15cm}
\caption{PointPillar FN examples with $\tau_{conf} = 0.3$. Red boxes are miss detections. Top: Missed car occluded. Bottom: Bicycle not detected even though clearly visible }
\label{fig:ppMiss03}
%\vspace{-0.25cm}
\end{figure}

In general it can be observed that using a threshold of 0.3 or lower, the majority of detection misses are occluded and should not cause a safety issue. Nevertheless the network missed to detect a clearly visible bicyclist, as shown in Figure \ref{fig:ppMiss03}. This also nicely underlines the motivation and need for a safety channel, to handle such corner cases of the main perception system.

\begin{figure}[t!]
\centering
\includegraphics[clip,trim=7.2cm 5.cm 8.cm 10.cm,width=0.95\columnwidth]{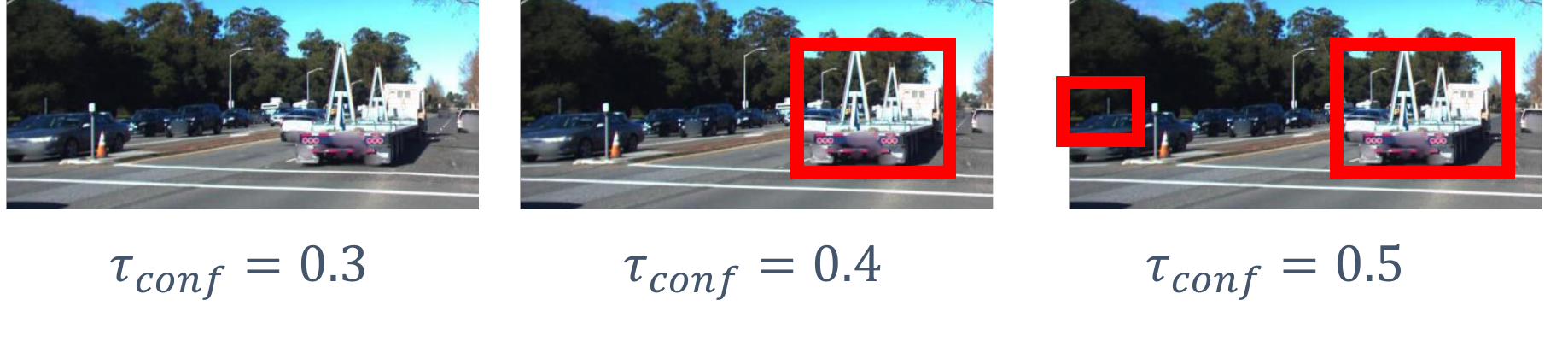}
\vspace{-0.15cm}
\caption{PointPillar results with different confidence thresholds. Red boxes are miss detections. Detection quality is very sensitive to $\tau_{conf}$}
\label{fig:ppMissThres}
\vspace{-0.45cm}
\end{figure}

When increasing the threshold also prominent objects are no longer reliably detected (see Figure \ref{fig:ppMissThres}). This highlights the sensitivity of the AI detection to parametrization. A small modification of 0.1 on $\tau_{conf}$ leads to a miss detection of an entire truck directly in front of the vehicle and therefore has huge impact to the vehicle safety. 

\subsection{Hierarchical Monitor Concept} 

In a second evaluation, we are estimating the quality of the created occupancy grid, which is the result of our proposed hierarchical monitor concept. The corresponding results are summarized in table~\ref{table:eval}-[1-4].
The occupancy grid should contain an occupied cell for regions that contain an object.
In order to evaluate the quality of the results we compare the cell regions with the bounding boxes of the annotations. Therefore, we calculate if there is an intersection of grid cell and the bounding box. 

As a metric we use the following:
\begin{description}
\item[FN] False Negative is a missed object. Therefore, an annotation in the ground truth that has no intersection with an occupied grid cell.
\item[TP] True Positive is an identified object. Therefore, an annotation that has an intersection with an occupied grid cell.
\item[FP] False Positive is a falsely occupied cell. So an occupied cell that has no intersection with an annotation. As the amount of reported FPs with this method is dependent on the used grid cell size - smaller grid cells, will create more false positives for same area -  we apply DBScan clustering to the FP cells. This will provide more reasonable results, as there will be only one FP report for adjacent FP cells.  
\end{description}

To get a better understanding about the contribution of the individual filters we evaluated several configurations.

\begin{description}
\item $\mathbf{HighConfidence_{sensitive}}$: In this configuration we use a very sensitive configuration of the height confidence filter (model-based probability filter). For the combination in Equation~\eqref{eq:combination} we assume $P_S$ and $P_N$ are 1. Therefore, all areas with points that have $P_O > 0.5$ will be considered as occupied. This configuration will provide the lower bound for FN and the upper bound of FP. So in the end the most conservative configuration with lowest possible Miss Rate and highest possible False Alarm Rate.\\   
\item  $\mathbf{HighConfidence}$ will be the default configuration. When evaluated standalone $P_S$ and $P_N$ are assumed to be 0. This gives highest possible Miss Rate and lowest possible False Alarm Rate \\
\item  \textbf{Objectiveness | Semantic Segmentation} will be combined according to Equation~\eqref{eq:combination}. \\
\end{description}

It can been seen that the experiments prove our assumptions. $\mathbf{HighConfidence_{sensitive}}$ shows the lowest amount of perception misses, with only 4 of all relevant objects being not detected by the system and therefore a recall of $99.87\%$ is achieved. In contrast to that, the $\mathbf{HighConfidence}$ filter shows with only 41 false alarms a high precision of $98.67\%$.
     
\subsubsection{False Negatives}

\begin{figure}
\includegraphics[clip,trim=0cm 4.cm 7.cm 0.cm,width=0.95\columnwidth]{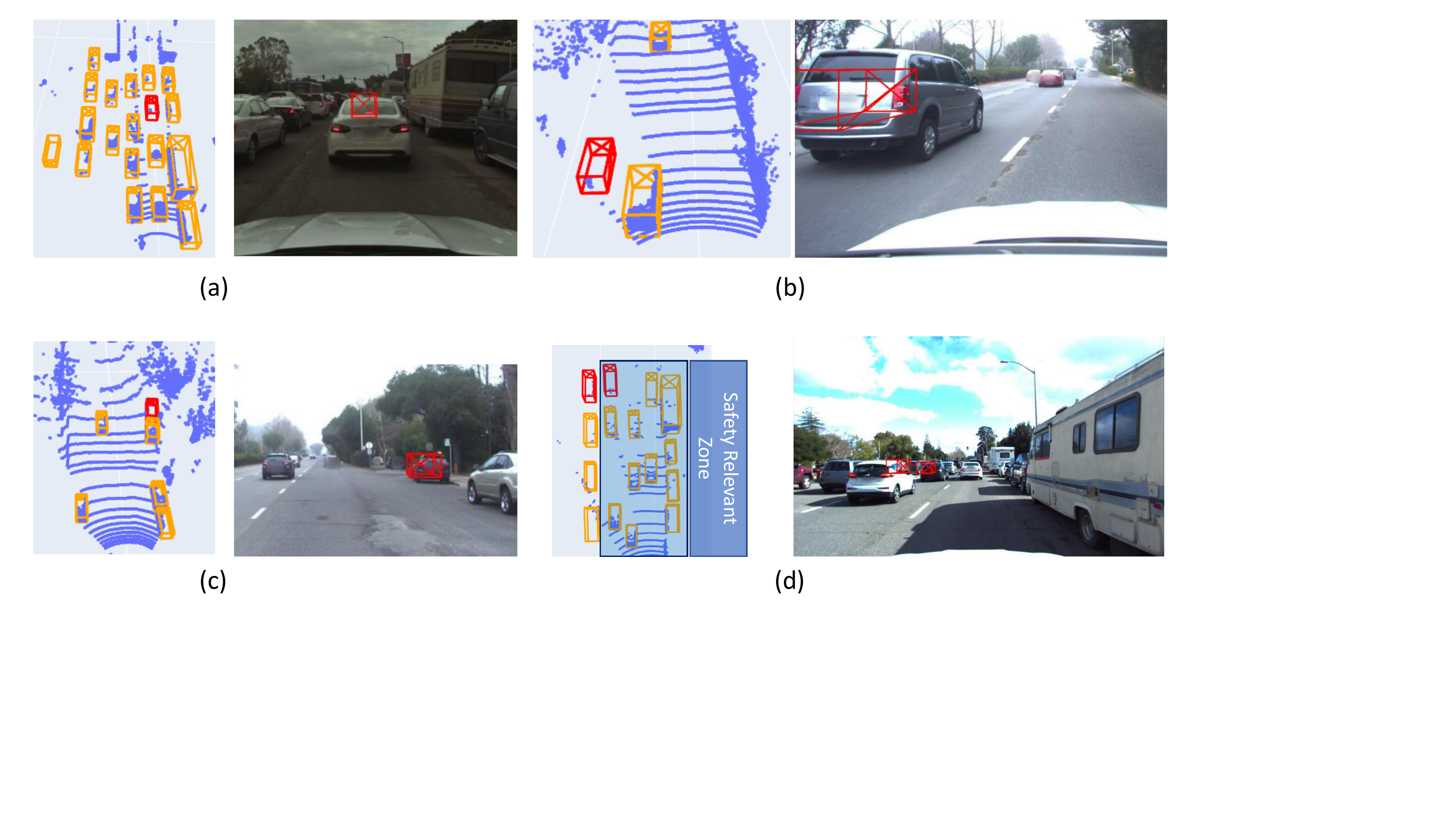}
\vspace{-0.15cm}
\caption{Red boxes missed detections of the monitor on the Lyft data set. All missed objects are heavily occluded and have no direct impact on the AV.}
\label{fig:lyftFN}
\vspace{-0.45cm}
\end{figure}

Figure~\ref{fig:lyftFN} shows all the missed detections on the evaluation. It can be seen that all missed objects are highly occluded. Also, there is no direct connection between the ego vehicle and the missed vehicles, as always at least one other car is in between. This closer vehicle has been successfully detected in all the situations. Therefore, it can be assumed that the missed vehicles have no safety impact on the ego vehicle.

\subsubsection{False Positives}

\begin{figure}
\includegraphics[clip,trim=2.5cm 8.cm 8.cm 0.cm,width=0.95\columnwidth]{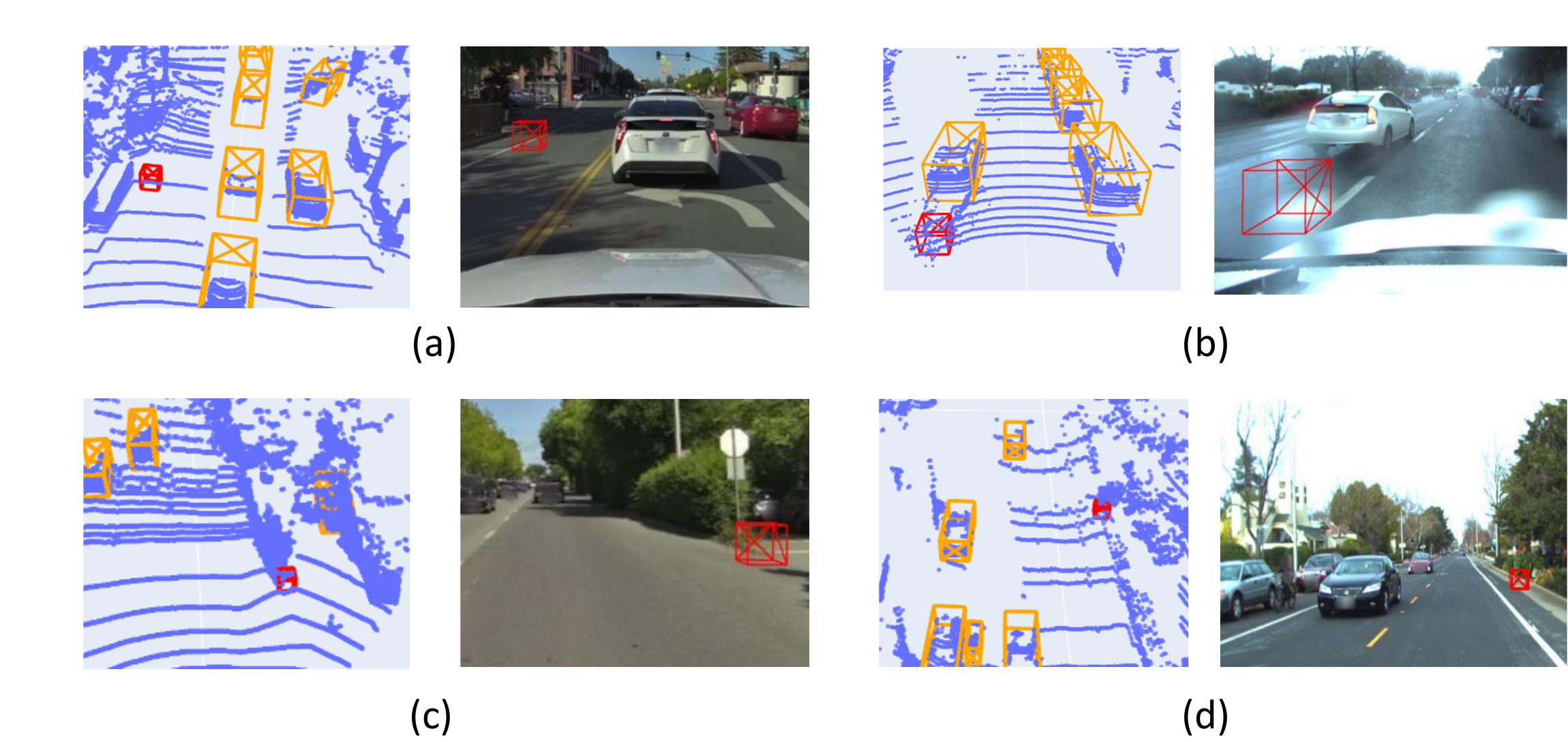}
\vspace{-0.15cm}
\caption{Red boundingbox: Exemplary False Positives of the monitor on the Lyft Dataset (a) Overhanging branches of a tree (b) Spray behind another car (c)\&(d) Vegetation close to road edge}
\label{fig:lyftFP}
\vspace{-0.45cm}
\end{figure}

Exemplary results of False Positives are shown in Figure~\ref{fig:lyftFP}. In our evaluation we have recognized several sources of False Positives, e.g. Figure~\ref{fig:lyftFP}(c - d). First of all there are obstacles close to the border of the road. This could be either vegetation that is reaching onto the road and therefore overlaps or on the sidewalk at the vicinity of the road. Here inaccuracies of the map lead to an in-proper separation and cause false positives.   

Another source of False Positives can be seen in  Figure~\ref{fig:lyftFP}(a). These are objects, especially trees, that reach over the road surface. If the branches of the trees are lower then the configured required height for traversal underneath, the system identifies those objects as obstacles.
It can be argued whether such an obstacle could have a safety impact or not. Nevertheless, it might be anyway advisable to avoid these kind of obstacles to do not risk damage on the vehicle or the sensors. It is also possible to use a different parameter setting to be less restrictive on the required height.

In Figure~\ref{fig:lyftFP}(b) the false positive is created by spray behind a vehicle on a wet road surface. These kind of false positives might be not too relevant, as they require a leading vehicle in front of the ego and would result in a slightly longer safety distance towards the leading car. Also it might be possible to adapt the filtering method to identify noise created by spray and therefore reduce the false positives in these situations. 

\subsubsection{Combination}

For the final monitor we want to combine the different filters according to Equation~\eqref{eq:combination}. Results of these combinations can be seen in table~\ref{table:eval}-[4]. As a baseline we use the $\mathbf{HighConfidence}$ filter. 
By the use of additional filters as complementary sources of information it is possible to reduce FNs down to the optimum of the $\mathbf{HighConfidence_{sensitive}}$. We also see that this comes at the cost of a small increase of FPs. The precision of the final combination drops down to $98.1 \%$ and is therefore $0.6 \%$ lower as the pure $\mathbf{HighConfidence}$ filter. 
This increase might be avoidable by the usage of additional filters, nevertheless the final results are low compared with results from PointPillars when considering comparable FN rates.

\subsection{Putting things together: Monitor + Primary Perception}

The last evaluation is the combination of the primary channel, i.e PointPillars, and the monitor, as described in Figure \ref{fig:MonitorIdea}.
Therefore, we feed the detections of the Point Pillar perception into our monitor. Here we used again the same $\tau_{conf}$ values as for the standalone evaluation of the PointPillars algorithm. 
The monitor compares the detected objects with the occupancy data as described in Section~\ref{realization:monitor}. The output of the monitor is a corrected object list, for which the results are described in Table~\ref{table:eval}-[10-14].

The evaluation shows that the combination with PointPillars does not lead to an increase of FNs. Hence, \textbf{all missed objects by PointPillars were successfully detected and flagged by the monitor}.
It can be also seen that the number of the FPs varies. 
If the primary perceptions has a high number of FPs as for the $\tau_{conf}=0.1$, the overall number FPs decreases from 224 to 179. Here the monitor helps to identify FPs detections. 
For the primary systems with higher Precision, the amount of FPs increases. 
But the amount of the final FPs is for all configurations smaller then the sum of the FPs of both channels. This is caused by the fact, that some of the FPs refer to the same location. These might be regions that contain measurments of real objects that are not part of the groundtruth or are FPs on both channels.

\begin{table*}[htbp]
\vspace{+0.25cm}
    %\resizebox{\textwidth}{!}{%
        \begin{tabularx}{ \textwidth}{@{}c@{}|c|cc|cccc|ccccc}
            \cmidrule{1-13}
            \multirow{2}{0.5in}{\thead{Setup \#}} & \multirow{2}{0.5in}{\thead{Data Set }}&  \multicolumn{2}{|c}{\thead{Primary Perception}} &  \multicolumn{4}{|c}{\thead{Filter}} & \multicolumn{5}{|c}{\thead{Results}}\\ 
             \cmidrule{3-13}
             & & \thead{PointPillars} &  \thead{$\tau_{conf}$} & \thead{$HC_{sensitive}$} & \thead{$HC$} & \thead{Objectiveness} & \thead{SemSeg} & \thead{FN} & \thead{FP} & \thead{TP}  & \thead{Precision} & \thead{Recall}\\ 
			\cmidrule{1-13}

            \cmidrule{1-13}
            1 & Lyft & & &\checkmark & & & &  4 & 539 & 3042 &   0.849483 &	0.998687 \\
            \cmidrule{1-13}
            2 & Lyft & && & \checkmark  & & &  15 & 41 & 3031 & 0.986654 &	0.995076  \\
            \cmidrule{1-13}
            3 & Lyft & && & \checkmark  & \checkmark  & &  9 & 41 & 3037 & 	0.986680 & 0.997045 	 \\          
            \cmidrule{1-13}
            4 & Lyft & & && \checkmark  & \checkmark  & \checkmark &  4 & 59 & 3042 & 0.980974 &	0.998687  \\          
            \cmidrule{1-13}
            
            \cmidrule{1-13}
            5 & Lyft & \checkmark &0.1& & & & &18 &224 &3028 & 0.931119	& 0.994091  \\
			%\cmidrule{1-13}
            6 & Lyft & \checkmark &0.2& & & & &20 &115 &3026 & 0.963387 &	0.993434  \\            
           % \cmidrule{1-13}
            7 & Lyft & \checkmark &0.3& & & & &34 &45 &3012 & 0.985280 &	0.988838 \\
            %\cmidrule{1-13}
            8 & Lyft & \checkmark &0.5& & & & &90 &9 &2956 & 0.996965 &	0.970453   \\
            %\cmidrule{1-13}		
            9 & Lyft & \checkmark &0.8& & & & &542 &0 &2504 & 1.0 &	0.822062   \\
            \cmidrule{1-13}	
            
  			\cmidrule{1-13}
            10 & Lyft & \checkmark &0.1& &\checkmark &\checkmark &\checkmark &4 &179 &3042 & 0.944427 & 0.998687 \\ 
             11 & Lyft & \checkmark &0.2& &\checkmark &\checkmark &\checkmark &4 &124 &3042 & 0.960834 &	0.998687 \\ 
            %\cmidrule{1-13}
			12 & Lyft & \checkmark &0.3& &\checkmark &\checkmark &\checkmark &4 &85 &3042 & 0.972817 &	0.998687   \\
			13 & Lyft & \checkmark &0.5& &\checkmark &\checkmark &\checkmark &4 &62 &3042 & 0.980026 &	0.998687   \\
			14 & Lyft & \checkmark &0.8& &\checkmark &\checkmark &\checkmark &4 &58 &3042 & 0.981290	& 0.998687   \\
            \cmidrule{1-13}
        \end{tabularx}

    %}
            \caption{Summary of the different evaluation runs [1-4] Evaluation of the standalone filter [5-9] Evaluation of PointPillars using different confidence thresholds $\tau_{conf}$ [10-14] Evaluation of the final system: PointPillars perception with different confidence thresholds in combination with the monitor }
	\label{table:eval}
\end{table*}

\section{Conclusion}
\label{sec:conclusion}

Perception systems of automated vehicles rely on AI-based detection algorithms. In general, these algorithms offer great detection quality, nevertheless they rely on the quality of the training data and thus there is always the possibility that objects are missed even though they are clearly visible, simply as these were not covered in any training data.

Consequently, it is hard to construct a proof that such an AI-based system is safe under all operating conditions. To address that gap, we presented a novel monitor to check the correctness of a given perception system. The monitor uses LiDAR information, and is realized as a combination of a rule-based probability filter, that allows the system to correctly identify prominent objects that may have been missed by the primary perception system, and additional filters that both increase detection rate and decrease false alarm rate.

We highlighted in our evaluation that a state-of-art AI detector for LiDAR perception can miss prominent and safety relevant objects in the surrounding of the vehicle. Also these detectors are sensitive to configuration parameters. A small change in the configuration of the confidence threshold can lead to miss detection of clearly visible objects. The use of lower confidence can reduce the miss detections, but only at the cost of more false alarms.

With our monitor architecture we are able to eliminate the majority of detection misses, and the few remaining ones are all heavy occluded and not in direct connection to the ego vehicle (thus not safety relevant). In addition to that the false alarm rate is still low ( $<2\%$).

Hence, in summary, with our approach we can improve safety of the LiDAR perception system and still maintain a high availability. It is worth mentioning, that in this work we focused on a single channel LiDAR system. In a final AV, it is desirable to use additional modalities in parallel, to gain robustness against sensor failures.

\section{Acknowledgments}
This research was funded by the German Ministry for Economic Affairs and Energy in the project SafeADArchitect (19A20013A).

\bibliographystyle{IEEEtran}
\bibliography{main}

\end{document}